\documentclass{article}

\PassOptionsToPackage{numbers, sort&compress}{natbib}
\usepackage[preprint]{neurips_2023}

\usepackage[utf8]{inputenc} % allow utf-8 input
\usepackage[T1]{fontenc}    % use 8-bit T1 fonts
\usepackage{hyperref}       % hyperlinks
\usepackage{url}            % simple URL typesetting
\usepackage{xurl}
\usepackage{booktabs}       % professional-quality tables
\usepackage{amsmath}
\usepackage{amssymb}
\usepackage{amsfonts}       % blackboard math symbols
\usepackage{nicefrac}       % compact symbols for 1/2, etc.
\usepackage{microtype}      % microtypography
\usepackage{xcolor}         % colors
\usepackage{epsfig}
\usepackage{graphicx}
\usepackage{multirow}
\usepackage{diagbox}
\usepackage{booktabs}
\usepackage[normalem]{ulem}
\useunder{\uline}{\ul}{}

\title{F?D: On understanding the role of deep feature spaces on face generation evaluation}
\author{%
  Krish Kabra \\
  Department of Electrical \& Computer Engineering\\
  Rice University\\
  Houston, TX 77005 \\
  \texttt{kk80@rice.edu} \\
  \And 
  Guha Balakrishnan \\
  Department of Electrical \& Computer Engineering\\
  Rice University\\
  Houston, TX 77005 \\
  \texttt{gb35@rice.edu} 
}

\begin{document}

\maketitle

\begin{abstract}
%The rapid advances of deep generative image models also necessitate improvements to the metrics used to evaluate them. 
Perceptual metrics, like the Fr\'echet Inception Distance (FID), are widely used to assess the similarity between synthetically generated and ground truth (real) images. The key idea behind these metrics is to compute errors in a deep feature space that captures perceptually and semantically rich image features. Despite their popularity, the effect that different deep features and their design choices have on a perceptual metric has not been well studied. In this work, we perform a causal analysis linking differences in semantic attributes and distortions between face image distributions to Fr\'echet distances (FD) using several popular deep feature spaces. A key component of our analysis is the creation of synthetic counterfactual faces using deep face generators. Our experiments show that the FD is heavily influenced by its feature space's training dataset and objective function. For example, FD using features extracted from ImageNet-trained models heavily emphasize hats over regions like the eyes and mouth. Moreover, FD using features from a face gender classifier emphasize hair length more than distances in an identity (recognition) feature space. Finally, we evaluate several popular face generation models across feature spaces and find that StyleGAN2 consistently ranks higher than other face generators, except with respect to identity (recognition) features. This suggests the need for considering multiple feature spaces when evaluating generative models and using feature spaces that are tuned to nuances of the domain of interest. 
\end{abstract}

% TLDR: We perform a causal analysis showing how different facial attributes and distortions to semantic regions impact the Fréchet distance across several popular deep image representations. 

\section{Introduction}
Rapid advances in generative image models such as variational autoencoders (VAEs) \cite{razaviGeneratingDiverseHighFidelity2019, vahdatNVAEDeepHierarchical2020}, generative adversarial networks (GANs) \cite{goodfellowGenerativeAdversarialNetworks2020, karrasStyleBasedGeneratorArchitecture2019a, karrasAnalyzingImprovingImage2020, chanEfficientGeometryAware3D2022}, and diffusion models \cite{hoDenoisingDiffusionProbabilistic2020, songMaximumLikelihoodTraining2021, rombachHighResolutionImageSynthesis2022, dhariwalDiffusionModelsBeat2021}, point to a future where synthetic images play a significant role in society \cite{guiReviewGenerativeAdversarial2023, wangReviewMedicalImaging2021, mcduff2023synthetic}. Therefore, it is crucial to continuously assess and improve how we evaluate the performances of these generative models~\cite{burnellRethinkReportingEvaluation2023}. In particular, synthesis evaluation metrics should capture several factors, including correlation to human perception, robustness to insignificant variations and noise, and sensitivity to domain-specific semantics. 

The gold standard in evaluating image generation quality is human annotation~\cite{zhouHYPEBenchmarkHuman2019}, which can provide nuanced and interpretable perceptual feedback, but comes at the cost of money and time. The current standards in automated evaluation are deep perceptual metrics like perceptual loss (LPIPS)~\cite{zhangUnreasonableEffectivenessDeep2018} and Fr\'echet Inception Distance (FID)~\cite{heuselGANsTrainedTwo2017}. These metrics embed images in the activation space of a deep neural network trained on a general auxiliary task and compute distances in that feature space. Deep perceptual metrics correlate better with human evaluation than classical metrics (e.g., PSNR)~\cite{zhangUnreasonableEffectivenessDeep2018, xuEmpircalStudyEvaluation2018}. Unfortunately, the complexity of deep feature spaces also makes them opaque and hard to interpret. For example, deep feature spaces will emphasize certain attributes over others (as we demonstrate in our experiments), and can even be influenced by spurious features unrelated to the domain of interest~\cite{kynkaanniemiRoleImageNetClasses2023}. Given that deep generative models are now typically competing with each other for less than 5 FID points, it is unclear what such differences mean semantically. When evaluating face generators, answering questions such as ``What effect does an imbalanced generation of skin tones have on FID?" or, ``What is the effect of consistent distortion of the eyes on FID?" is crucial in helping engineers better understand their evaluation metric, which ultimately will enable them to mitigate biases inherent in their models and/or improve generation quality. % and ultimately improve their generative models or, and mitigate biases of their generative models. 

In this work, we propose a strategy to causally evaluate the effect variations in domain-specific characteristics have on a perceptual evaluation metric using synthetic data. We focus on face generation, the most popular domain for image synthesis studies with many important societal implications in applications like face analysis/recognition \cite{balakrishnanCausalBenchmarkingBiasin2021}, deepfakes~\cite{TolosanaDeepfakesBeyond2020}, virtual avatars~\cite{pinkney2020resolution, abdal20233davatargan}, and even healthcare \cite{wangSyntheticGenerationFace2022}. We consider two types of face manipulations: semantic attributes (e.g., hats, skin tone, hair length) and distortions (blur). We perform causal studies, using experimental interventions that manipulate a single characteristic of an image at a time, and measure their effects on an evaluation metric (in our experiments, Fr\'echet distance) across different image spaces. For semantic interventions, we use deep face generators to construct a dataset of synthetic face pairs that differ (approximately) by only a single feature of interest. For distortions, we apply blur to semantic facial regions inferred by a face segmentation model. 

Using this synthetic data, we evaluated Fr\'echet distances (FD) in six deep feature spaces trained on both general-domain (ImageNet \cite{dengImageNetLargescaleHierarchical2009}, CLIP \cite{radfordLearningTransferableVisual2021}) and in-domain (face) datasets, and with fully-supervised, semi-supervised, and unsupervised objective functions. We find that the sensitivity of FD to semantic attributes depends heavily on the training dataset \emph{and} objective function used to train the feature space. For example, feature spaces trained on ImageNet tend to overly emphasize accessories like hats and eyeglasses, while ignoring important facial semantics such as eyes, expression, geometry, and skin texture. Moreover, feature spaces trained on in-domain face datasets with contrastive learning objectives emphasize features of the skin (e.g. skin tone and texture) but ignore details related to hair and background. Finally, we evaluated popular deep generative face models using FD, precision, and recall \cite{kynkaanniemiImprovedPrecisionRecall2019} metrics computed over these feature spaces. Our findings show that while StyleGAN2 outperforms other generators on most feature spaces, in a face identity (recognition) feature space it has significantly worse recall and FD than a popular diffusion model (LDM), and worse precision than a 3D-aware GAN (EG3D). 

The results of this study demonstrate that deep feature spaces have significant and unique biases over in-domain attributes due to both training data and objective functions. These biases should be understood by researchers during synthesis evaluation. In addition, our experiments with face generators demonstrate the importance of considering multiple feature spaces during evaluation, and particularly those tuned to crucial details for the domain of interest.

\subsection{Related Work \& Background}
%Classically, this can be achieved by computing the average log-likelihood of generative models. However, due to the high-dimensionality of images, it can be the case that a model can achieve high likelihood and poor sample quality, or poor likelihood and great sample quality~\cite{theisNoteEvaluationGenerative2016}. Instead, 
Deep generative model evaluation involves computing the similarity between generated and real image distributions. Metrics such as the Inception Score (IS)~\cite{salimansImprovedTechniquesTraining2016} and Fr\'echet Inception Distance (FID)~\cite{heuselGANsTrainedTwo2017} embed images into a lower-dimensional perceptual feature space derived from the final layers of a deep convolutional neural network (InceptionV3~\cite{szegedyRethinkingInceptionArchitecture2016}) trained on ImageNet~\cite{dengImageNetLargescaleHierarchical2009}. 

FID is currently the \textit{de facto} image generation evaluation metric. It assumes that two Inception-embedded image distributions are multivariate Gaussian, and computes the Fr\'echet distance (FD), otherwise known as 2-Wasserstein or Earth Mover's distance \cite{dowsonFrechetDistanceMultivariate1982}: 
\begin{equation}
\label{eq:wasserstein}
    \text{FD}(\mathbf{\mu_1}, \mathbf{\Sigma_1}, \mathbf{\mu_2}, \mathbf{\Sigma_2}) = ||\mathbf{\mu_1} - \mathbf{\mu_2}||_2^2 + \text{Tr} \Big( \mathbf{\Sigma_1} + \mathbf{\Sigma_2} -2 (\mathbf{\Sigma_1}\mathbf{\Sigma_2})^\frac{1}{2} \Big),
\end{equation}
where $(\mathbf{\mu_1}, \mathbf{\Sigma_1})$ and $(\mathbf{\mu_2}, \mathbf{\Sigma_2})$ are the sample means and covariances of the image set embeddings (i.e., real and generated images), and $\text{Tr}(\cdot)$ is the matrix trace. 

Several recent works have exposed limitations of FD and suggest alternate evaluation metrics~\cite{borjiProsConsGAN2022}. One major drawback of FD is its high bias, requiring $50,000$ samples or more to yield accurate estimates. To combat this, Bi\'nkowski \textit{et al.}~\cite{binkowskiDemystifyingMMDGANs2023} propose relaxing the Gaussian assumption using a polynomial kernel and computing the squared maximum mean discrepancy, whereas Chong \textit{et al.}~\cite{chongEffectivelyUnbiasedFID2020} propose estimating a bias-free FD by extrapolating the score using the fact that it is linear with respect to $\frac{1}{N}$. Another limitation of FD is its obfuscation of sample quality and variation. To circumvent this, Sajjadi \textit{et al.}~\cite{sajjadiAssessingGenerativeModels2018} and Kynk\"{a}\"{a}nniemi \textit{et al.}~\cite{kynkaanniemiImprovedPrecisionRecall2019} propose separating these two components of evaluation into separate metrics: \emph{precision}, which quantifies sample quality, and \emph{recall}, which quantifies sample coverage. 

More recently, FID has been shown to be limited by unwanted biases and sensitivities. Parmar \textit{et al.}~\cite{parmarAliasedResizingSurprising2022} show that FID is sensitive to imperceptible artifacts caused by image processing operations such as resizing, compression and quantization. Kynk\"{a}\"{a}nniemi \textit{et al.}~\cite{kynkaanniemiRoleImageNetClasses2023} show that out-of-domain (ImageNet) features in the background of face images also strongly affect FID. Morozov \textit{et al.}~\cite{morozovSelfSupervisedImageRepresentations2021} advocate computing FD using self-supervised feature spaces, arguing that these features are more transferable and robust than those in Inception. 

Our work is inspired by these studies and builds on them in several ways. First, our focus is a \emph{causal} understanding of how \emph{in-domain} facial semantics affect FD in several deep feature spaces. Second, we propose a novel strategy to perform this evaluation via synthesizing counterfactual image pairs with deep generative models. Third, we focus on one important domain (faces), which allows us to perform more nuanced semantic evaluations (e.g., how skin tone or nose distortions affects FD) than the general case considered by all previous studies. Finally, we also provide evaluations of several popular deep face generation models in these feature spaces, yielding insight into how these models compare to one another.

\begin{figure}[t!] 
    \centering
    \includegraphics[width=\linewidth]{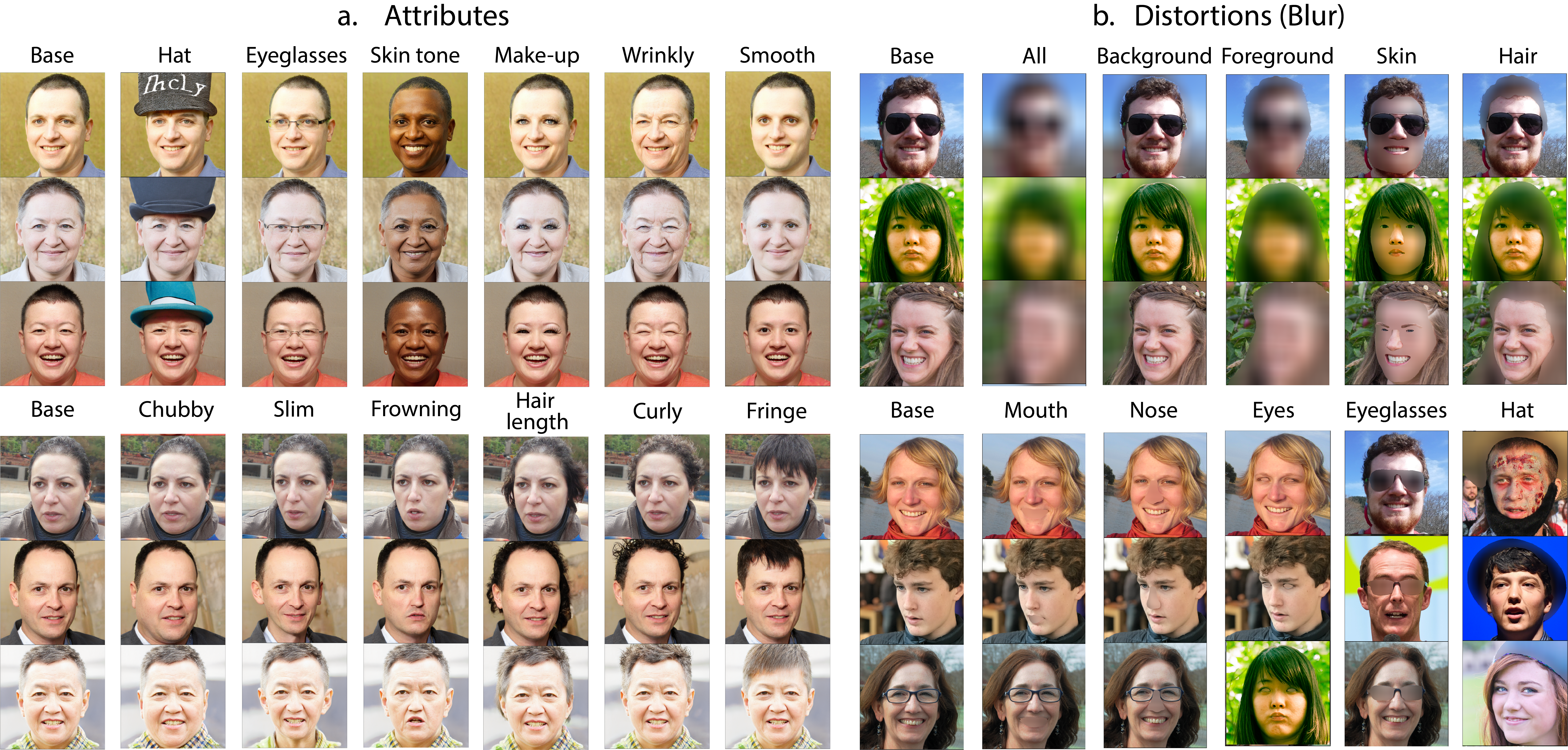}
    \caption{\textbf{Sample images in our proposed counterfactual face dataset}. (a) The first column shows synthetically generated base faces with predefined ``neutral'' characteristics. We manipulate each base face along different attribute of interest, shown in the remaining columns. (b) The first column (base) shows real faces from FFHQ~\cite{karrasStyleBasedGeneratorArchitecture2019a}. Faces in the remaining columns are distorted versions of the base faces with blur added to specific semantic regions. Note that we show different faces for some of the eyes, eyeglasses, and hat examples because the corresponding base faces do not contain those regions (eyeglasses are considered to occlude and therefore remove the eye regions).}
    \label{fig:counterfactual-dataset}
\end{figure}

% ----------------------------------------------------------- %
\section{Methods} \label{sec:methods}

For a given deep feature space, our goal is to quantify the sensitivity of an evaluation metric to image characteristics. In our experiments, we focus on face images and FD, and so we describe our methods here in that context. We form two questions for a given feature space: 
\begin{enumerate}
    \item How do differences between the semantic attribute distributions of two face image sets quantitatively affect FD?
    \item How do distortions localized to a semantic region of a face quantitatively affect FD? 
\end{enumerate}

These questions align with two broad image characteristics that a generative model must capture: (1) semantic attributes for the domain, and (2) realistic details. Answering these questions requires causal reasoning, and ideally a \emph{counterfactual} dataset consisting of pairs of faces that are identical except for a difference along one characteristic (i.e., semantic attribute or distortion). Real-face datasets contain significant attribute correlations~\cite{balakrishnanCausalBenchmarkingBiasin2021} and are therefore not appropriate. Instead, we propose a synthetic approach. In the following sections, we outline the proposed methodology to construct synthetic data to answer each question\footnote{We will make our synthesized dataset publicly available.}. Example images synthesized by the proposed methods are shown in Figure~\ref{fig:counterfactual-dataset}.

% In the following sections, we outline our methodology to synthesize such data. %In our experiments, we focus on FD metric. 

\subsection{Measuring the effect of semantic attribute differences on Fr\'echet distance}
\label{sec:method-attributes}
\begin{figure}[t!] 
    \centering
    \includegraphics[width=\linewidth]{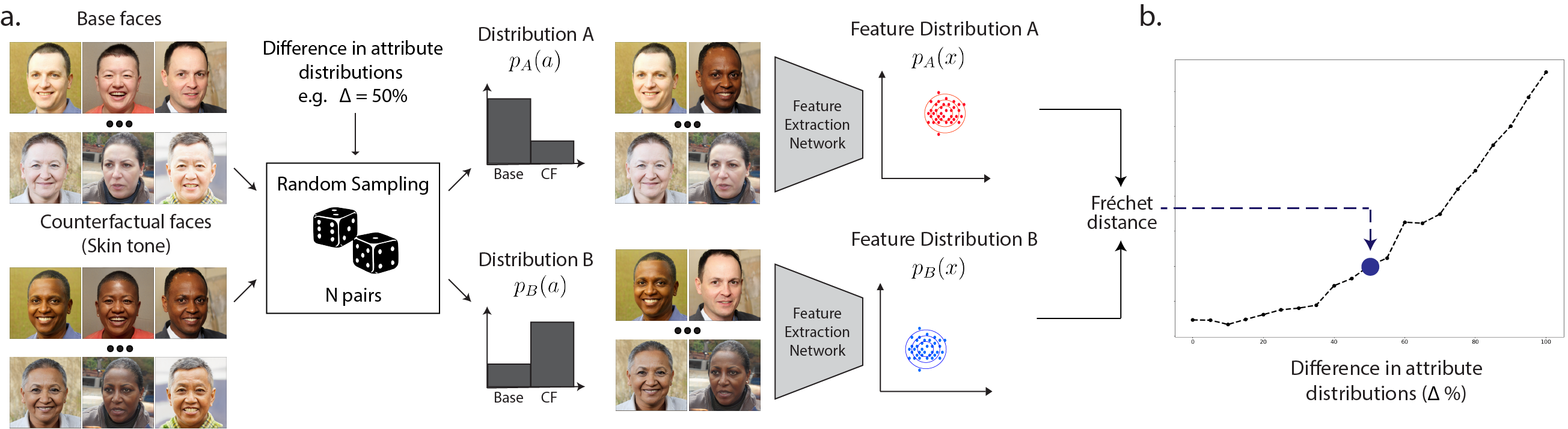}
    \caption{\textbf{Overview of our method for causally assessing the sensitivity of Fr\'echet distance to facial attributes.} a. We generate a set of base and counterfactual (CF) face pairs for an attribute or semantic distortion. In this figure, we use skin tone as an example. Given a difference in proportion for this attribute between two distributions ($\Delta \in [0\%, 100\%]$) and the number of faces per set ($N$), we construct two image sets A and B by randomly assigning base and CF faces to them such that this difference is achieved. We then extract the features for each image set using a pre-trained deep model (e.g., Inception, CLIP), and compute the FD between the two feature distributions. b. By creating set pairs for $\Delta \in [0\%, 100\%]$, we can generate a curve that summarizes the causal effect of a difference in attribute proportions on FD computed in a feature space.}
    \label{fig:analysis-overview}
\end{figure}

\begin{figure}[t!] 
  \centering
  \includegraphics[width=\linewidth]{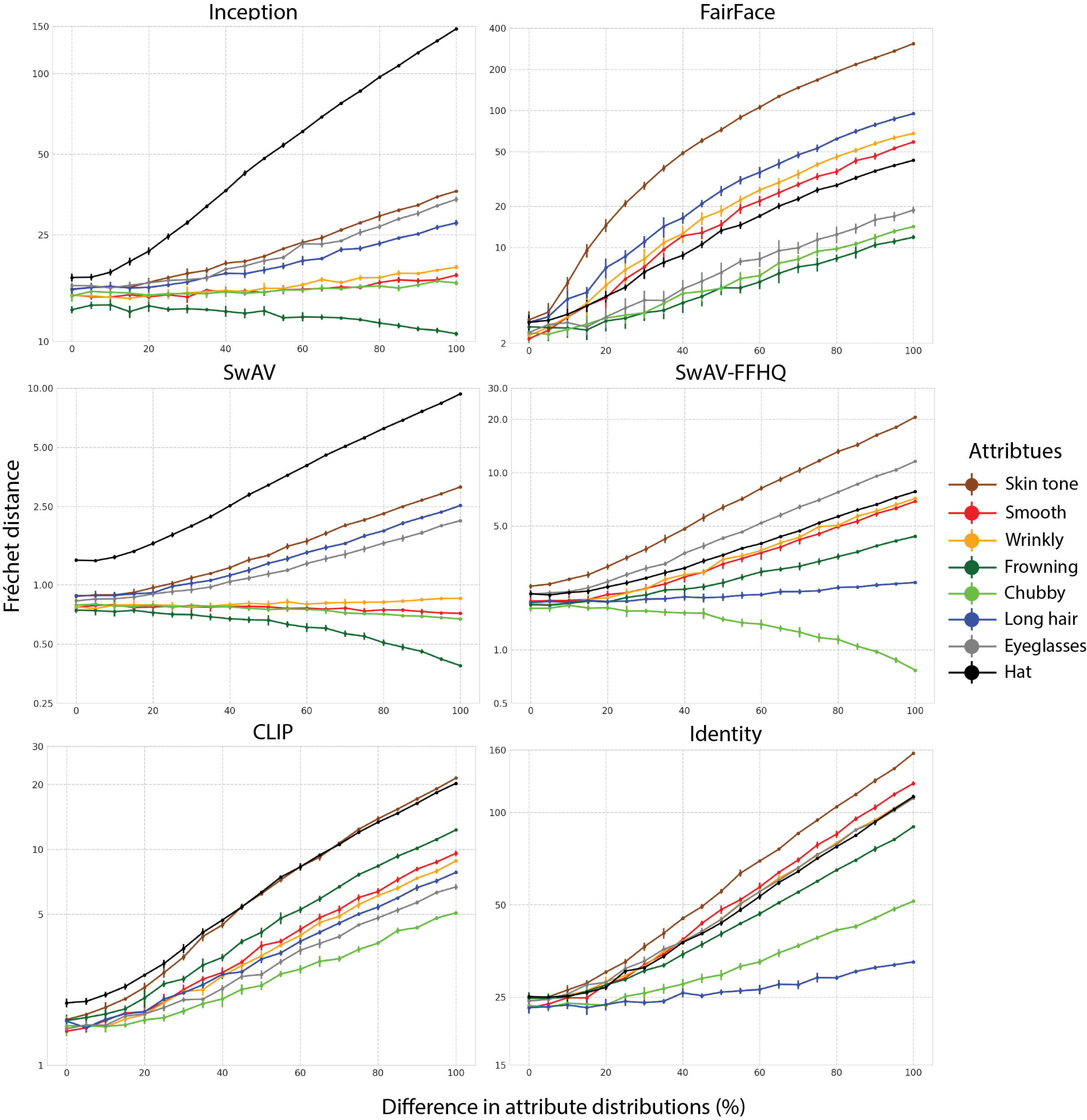}
  \caption{\textbf{Results for causal sensitivity analysis of Fr\'echet distances (FD) in different feature spaces with respect to semantic attributes.} For each percentage difference in attribute proportion (i.e. point along the x-axis), we sample 10 random draws of 1000 counterfactual face pairs to construct face sets, from which the FD in a feature space is computed, shown by the y-axis in log scale. See Figure~\ref{fig:analysis-overview} and Section~\ref{sec:method-attributes} for a more detailed description of the approach. The three feature spaces on the left are trained on general-domain data, while the ones on the right are trained on in-domain (face) data. Each feature space under- or over-emphasizes certain attributes based on its training dataset and objective functions. Note that absolute values of FD are not meaningful to compare across feature spaces due to arbitrary scaling differences.}
  \label{fig:causal-analysis-attrs}
\end{figure}

Consider two image sets $A$ and $B$ with feature distributions $p_A(x)$ and $p_B(x)$, where $x \in R^D$ is a feature space of an image. Furthermore, assume that $A$ and $B$ are identical except for a difference in their distributions over one semantic binary attribute with value $a\in\{0,1\}$, which we denote by $p_A(a)$ and $p_B(a)$. Our goal is to quantify how the difference in attribute proportions between $p_A(a)$ and $p_B(a)$ (ranging from $0\%$ when identical to $100\%$ when completely dissimilar), affects $\text{FD}(\mu_A, \Sigma_A, \mu_B, \Sigma_B)$, the FD between $p_A(x)$ and $p_B(x)$. Figure~\ref{fig:analysis-overview} describes our analysis methodology to do so. We construct multiple sets of nearly identical faces using deep generative models (described below), each consisting of different proportions of values to $a$, and compute FD between the pairs to yield a curve summarizing causal effects (see Figure~\ref{fig:analysis-overview}-right, and Figure~\ref{fig:causal-analysis-attrs}).

This analysis requires the creation of sets $A$ and $B$, counterfactual face sets that differ based on only $a$. We use a two-step process to create this data synthetically. First, we synthesize a set of \emph{base faces} that exhibit predefined uniform characteristics of light skin tones and short hair, and no: facial hair, make-up, frowning expressions, hats, or eyeglasses. To do this, we obtained the face generation models of a previous facial causal benchmarking study~\cite{balakrishnanCausalBenchmarkingBiasin2021} based on StyleGAN2~\cite{karrasAnalyzingImprovingImage2020} trained on the Flickr-Faces-HQ (FFHQ) dataset~\cite{karrasStyleBasedGeneratorArchitecture2019a} and orthogonalized linear latent space traversals (OLLT).%\footnote{The hyperplane coefficients for age, gender, hair length, and skin tone attributes were graciously provided by the authors upon request.}. 
We filter these faces via human evaluations to ensure they meet the defined criteria. In our experiments, we used a total of $1427$ filtered base faces. 

In the second step, we synthesize counterfactual pairs from the base faces for each attribute $a$. In our experiments, we analyzed 12 binary attributes corresponding to various facial semantics including geometry, skin tone, skin texture, hairstyle, and accessories. The attributes analyzed are shown by the columns in Figure~\ref{fig:counterfactual-dataset}a. We utilize one of three different image manipulation methods based on the attribute type: (1) OLLT, (2) StyleCLIP~\cite{patashnikStyleCLIPTextDrivenManipulation2021}, and (3) image inpainting with Stable Diffusion~\cite{vonplatenDiffusersStateoftheartDiffusion2023}. We choose the best method for each attribute based on a qualitative assessment of how well each method can manipulate the attribute while holding others constant. We show some example attribute counterfactuals in Figure~\ref{fig:counterfactual-dataset}a. We provide a complete account of models, experimental parameters, and details used to create the synthetic dataset in Supplementary. 

\subsection{Measuring the effect of blurring semantic regions on Fr\'echet distance}
\label{sec:method-blur}
\begin{figure}[t!] 
  \centering
  \includegraphics[width=\linewidth]{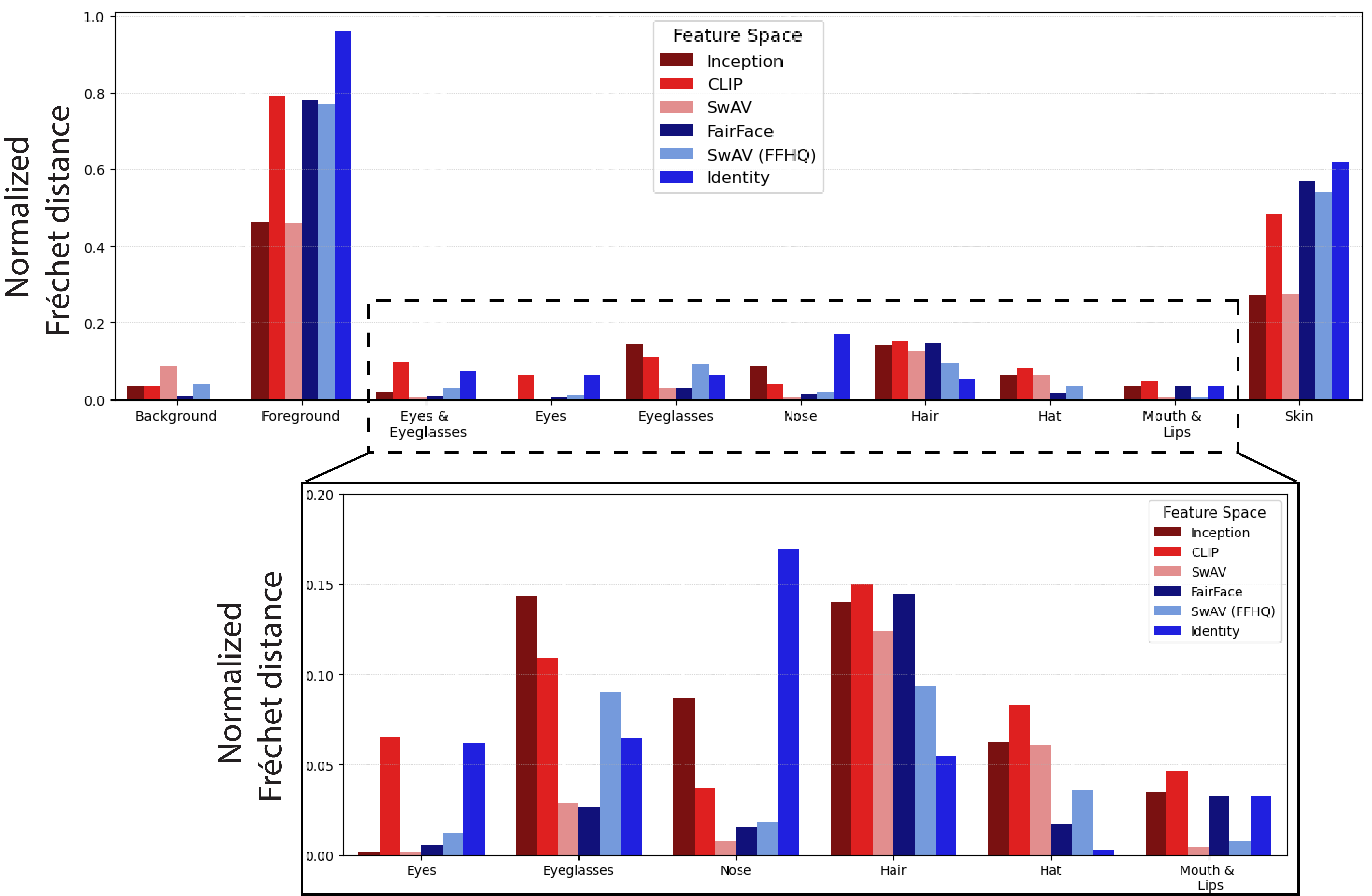}
  \caption{\textbf{Results for semantic region distortion (blur) analysis.} (Top) Bar plot comparing the normalized Fr\'echet distances per semantic region blur. We normalize distances in each feature space by dividing by the distance between the original and fully blurred image set (``All'' in Figure~\ref{fig:causal-analysis-attrs}) in that space. (Bottom) Zoomed-in plot to clearly visualize results for semantic regions that occupy less than 25\% of the face image on average.} 
  \label{fig:causal-analysis-distortions2}
\end{figure}

The purpose of this analysis is to understand how a systematic distortion outputted by a face generator for a specific semantic region (e.g., nose, hair) impacts FD. In our experiments, we focused on heavy blur (see Figure~\ref{fig:counterfactual-dataset}, though many others may also be explored~\cite{zhangUnreasonableEffectivenessDeep2018}. For each region, we use real FFHQ face images that contain that region (accessories like hats and eyeglasses are not in every image) as one distribution (set $A$), and apply Gaussian blur to these images \emph{only in that region} using segmentation masks obtained from a public face segmentation model\footnote{https://github.com/zllrunning/face-parsing.PyTorch} (set $B$). %with kernel size of $111\times111$ pixels and standard deviation $\sigma = 100$ pixels to create set $B$. To localize the distortions, we replace the areas of the original image corresponding to the facial part being examined with the blurred face  
We considered 9 regions in our experiments, as shown in Figure~\ref{fig:counterfactual-dataset}b. For our analysis, we simply report FD with respect to distorting each semantic region (see Figure~\ref{fig:causal-analysis-distortions2}).
% ----------------------------------------------------------- %

\section{Experiments}
\label{sec:experiments}
We conduct our analyses using six deep feature spaces with publicly available parameters (a list of model and parameter URLs are in Supplementary):
\begin{enumerate}
    \item \textbf{Inception}: V3 model trained on the ILSVRC-2012 (ImageNet) dataset for classification~\cite{szegedyRethinkingInceptionArchitecture2016}.
    \item \textbf{CLIP}: ViT-B/32 model trained on OpenAI's proprietary image-text pair dataset using a contrastive loss~\cite{radfordLearningTransferableVisual2021}.
    \item \textbf{SwAV}: ResNet-50 model trained on the ILSVRC-2012 (ImageNet) for unsupervised learning of visual features by contrasting cluster assignments~\cite{caronUnsupervisedLearningVisual2020}.
    \item \textbf{FairFace}: ResNet-34 model trained on FairFace for race, age, and gender classification~\cite{karkkainenFairFaceFaceAttribute2021}.
    \item \textbf{SwAV-FFHQ}: ResNet-50 model trained on FFHQ for unsupervised learning of facial features by contrasting cluster assignments~\cite{caronUnsupervisedLearningVisual2020, baranchukLabelEfficientSemanticSegmentation2022}.
    \item \textbf{Identity}: ResNet-34 model trained on Glint360k for facial recognition with a contrastive loss~\cite{dengArcFaceAdditiveAngular2019}.
\end{enumerate}

The first three feature spaces are trained on general-domain (non-face) data, while the last three are trained on faces only.

\subsection{Causal sensitivity of feature spaces to image characteristics}
We first present sensitivity analyses of FD with respect to facial attribute proportions. Figure~\ref{fig:causal-analysis-attrs} presents our results. Our total dataset used for this analysis contains 1427 counterfactual face pairs. For each percentage difference in attribute proportion (i.e. point along the x-axis), we sample 10 random draws of 1000 pairs to construct face sets, from which the FD in a feature space is computed. For clarity, we plotted only 8 attributes (full results for all 12 attributes are in Supplementary). The points and error bars shown in the plots correspond to the mean and standard deviation respectively. A direct comparison of FD values across feature spaces is not meaningful, as the scale of distances vary across features. However, the difference in trends between the attribute curves may be compared across plots. For example, Inception and SwAV clearly emphasize hats with respect to other feature spaces, while FairFace and SwAV-FFHQ emphasize skin tone.  

Next, we present sensitivity analyses of FD with respect to localized distortions. In order to compare FD across different feature spaces, we normalize distances by dividing them by the distance between the original and entirely blurry images (``all" category in Figure \ref{fig:counterfactual-dataset}) in that feature space. We show unnormalized FD for each feature space in Supplementary. There are significant differences in how each feature space is affected by regional distortions. For example, Inception and SwAV are virtually unaffected by the eyes compared to other spaces, and Identity is most affected by the nose.

\subsection{Analysis of face generators in different feature spaces}

\begin{table}[t!]
\caption{\textbf{Generative model evaluation using different deep image spaces and metrics.} We evaluate 50K images synthesized by each generative models with respect to the full FFHQ (70K) dataset. For each feature space, we highlight the top three performing models with the following key: {\ul\textbf{First}}, \textbf{Second}, {\ul Third}. Note that FD values are not meaningful to compare across feature spaces due to arbitrary scaling differences. StyleGAN2 generally outperforms all other models, but in Identity space is worse in Fr\'echet distance and Recall than LDM, and worse in Precision than EG3D.}
\label{tab:gen-model-eval}
\centering

Fr\'echet Distance ($\downarrow$)
\begin{tabular}{lcccccc}
    \toprule
    \multirow{2}{*}{} & \multirow{2}{*}{Inception} & \multirow{2}{*}{CLIP} & \multirow{2}{*}{SwAV} & \multirow{2}{*}{FairFace} & SwAV & \multirow{2}{*}{Identity} \\
     &  &  &  &  & \multicolumn{1}{l}{(FFHQ)} &  \\
     \midrule
    StyleGAN2~\cite{karrasAnalyzingImprovingImage2020} & {\ul \textbf{3.1}} & {\ul \textbf{1.8}} & {\ul \textbf{0.6}} & {\ul \textbf{1.6}} & {\ul \textbf{0.4}} & \textbf{17.8} \\
    StyleGAN2 (Truncated) & 21.0 & 8.2 & {\ul 2.0} & {\ul 27.1} & {\ul 4.0} & 61.2 \\
    EG3D~\cite{chanEfficientGeometryAware3D2022} & {\ul 16.5} & {\ul 7.0} & 2.1 & 34.2 & 9.2 & 162.0 \\
    EG3D (Truncated) & 40.2 & 13.0 & 3.3 & 41.8 & 16.7 & 221.3 \\
    LDM~\cite{rombachHighResolutionImageSynthesis2022} & \textbf{10.0} & \textbf{3.6} & \textbf{1.7} & \textbf{10.9} & \textbf{1.4} & {\ul \textbf{6.9}} \\
    NVAE~\cite{vahdatNVAEDeepHierarchical2020} & 35.9 & 9.7 & 5.4 & 56.9 & 5.8 & {\ul 44.1} \\
    \bottomrule \\
\end{tabular}

Precision (\%) ($\uparrow$)

\begin{tabular}{lcccccc}
    \toprule
    \multirow{2}{*}{} & \multirow{2}{*}{Inception} & \multirow{2}{*}{CLIP} & \multirow{2}{*}{SwAV} & \multirow{2}{*}{FairFace} & SwAV & \multirow{2}{*}{Identity} \\
     &  &  &  &  & \multicolumn{1}{l}{(FFHQ)} &  \\
     \midrule
    StyleGAN2 & 67.4 & {\ul 77.0} & \textbf{79.1} & {\ul 84.5} & \textbf{74.3} & 59.4 \\
    StyleGAN2 (Truncated) & {\ul \textbf{83.3}} & {\ul \textbf{89.0}} & {\ul \textbf{89.8}} & {\ul \textbf{88.7}} & {\ul 67.7} & \textbf{88.0} \\
    EG3D & 67.1 & 61.7 & 55.3 & 63.2 & 48.5 & {\ul 86.0} \\
    EG3D (Truncated) & \textbf{79.8} & \textbf{82.8} & 72.1 & 71.1 & 38.6 & {\ul \textbf{92.8}} \\
    LDM & {\ul 72.2} & 72.0 & {\ul 74.7} & \textbf{85.6} & {\ul \textbf{78.8}} & 37.8 \\
    NVAE & 65.3 & 57.7 & 69.5 & 82.3 & 49.0 & 65.5 \\
    \bottomrule \\
\end{tabular}

Recall (\%) ($\uparrow$)

\begin{tabular}{lcccccc}
    \toprule
    \multirow{2}{*}{} & \multirow{2}{*}{Inception} & \multirow{2}{*}{CLIP} & \multirow{2}{*}{SwAV} & \multirow{2}{*}{FairFace} & SwAV & \multirow{2}{*}{Identity} \\
     &  &  &  &  & \multicolumn{1}{l}{(FFHQ)} &  \\
     \midrule
    StyleGAN2 & {\ul \textbf{50.2}} & {\ul \textbf{42.3}} & {\ul \textbf{25.1}} & {\ul \textbf{81.9}} & {\ul \textbf{79.5}} & \textbf{3.5} \\
    StyleGAN2 (Truncated) & 26.5 & 14.8 & 7.1 & 61.1 & {\ul 66.3} & 0.7 \\
    EG3D & {\ul 26.9} & {\ul 20.3} & {\ul 9.7} & \textbf{80.7} & 20.9 & 0.0 \\
    EG3D (Truncated) & 11.1 & 5.6 & 1.7 & 49.8 & 11.9 & 0.0 \\
    LDM & \textbf{38.5} & \textbf{38.5} & \textbf{10.6} & {\ul 77.8} & \textbf{71.3} & {\ul \textbf{22.8}} \\
    NVAE & 12.1 & 10.4 & 0.4 & 55.7 & 46.5 & {\ul 1.9} \\
    \bottomrule 
\end{tabular}
\end{table}

Next, we present evaluations of four popular, publicly available face generation models using metrics computed in each feature space: StyleGAN2~\cite{karrasAnalyzingImprovingImage2020}, EG3D~\cite{chanEfficientGeometryAware3D2022}, latent diffusion model (LDM)~\cite{rombachHighResolutionImageSynthesis2022}, and Nouveau variational autoencoder (NVAE)~\cite{vahdatNVAEDeepHierarchical2020}. For StyleGAN2 and EG3D, we evaluate the models both with and without truncation~\cite{marchesi2017megapixel, brock2018large} ($\psi = 0.7$, truncation cutoff $= 14$). We evaluate models using FD and $k-$nearest neighbors precision and recall metrics~ \cite{kynkaanniemiImprovedPrecisionRecall2019}. These precision and recall measures approximate sample quality (realism) and sample coverage, respectively. We use the entire FFHQ dataset ($70,000$ images) and $50,000$ samples from each generative model. Complete results are shown in Table \ref{tab:gen-model-eval}.

\section{Discussion}
\label{sec:discussion}
In this section, we discuss key observations from our experiments, followed by limitations.

\textbf{Feature spaces learned using ImageNet under-emphasize important facial semantics regardless of the training objective.} Figure~\ref{fig:causal-analysis-attrs} illustrates that FD in feature spaces learned using ImageNet (Inception and SwAV) are most sensitive to differences in the proportion of hats, consistent with findings from Kynk\"{a}\"{a}nniemi \textit{et al.} \cite{kynkaanniemiRoleImageNetClasses2023}. However, interestingly, the FD computed using SwAV features are also sensitive to hats, even though that model is not explicitly trained to classify ImageNet classes. This is reasonable since self-supervised learning is known to be an effective pretraining strategy for ImageNet classification. 

The plots also illustrate that FD computed using ImageNet-learned spaces are highly insensitive to distributional differences in skin texture (``wrinkly" and ``smooth"),  geometry (``chubby"),  and expression (``frowning"). %This is an alarming observation, suggesting these ImageNet-based FD are highly insensitive to these important facial semantics. 
This result is also affected by a subtle interplay between the mean and trace terms of the FD in Equation~\eqref{eq:wasserstein}(see Supplementary). As the two distributions become more skewed in our sensitivity analyses (towards $0$ or $100$ \% in Figure~\ref{fig:causal-analysis-attrs}), the distribution means become more dissimilar, but their variances also decrease and reduce the trace term. This suggests another challenge in using FD alone: they can obfuscate differences in distribution modes versus distribution shapes. % suggesting another difficulty in diagnosing the root cause of variations using FD alone. %, which is to say that FD obfuscates precision and recall \cite{sajjadiAssessingGenerativeModels2018, kynkaanniemiImprovedPrecisionRecall2019}.  

Figure~\ref{fig:causal-analysis-distortions2} shows that FD computed using Inception and SwAV spaces are insensitive to the blurring of the eyes, and SwAV is insensitive to the blurring of the nose and mouth. This shows that systematic degradations to the eyes, nose, or mouth, will not impact the FD in ImageNet-based feature spaces. Generative model designers should pay extra attention to these semantic ``blind spots.''   

\textbf{The training objective influences which facial semantics are emphasized by a deep feature space.} Figure~\ref{fig:causal-analysis-attrs} shows that while in-domain feature spaces (FairFace, SwAV-FFHQ, Identity) are all highly sensitive to differences in skin tone, skin texture, and facial accessories, there do exist several notable dissimilarities. For example, FairFace is far more sensitive to hair length, compared to SwAV-FFHQ and Identity. This is further supported by the relatively small effect that blurring the hair has on SwAV-FFHQ and Identity compared to FairFace. Another notable distinction is that both FairFace and SwAV-FFHQ fail to capture distortions localized to the eyes, nose, mouth, and lips, whereas Identity does.  
We speculate that these differences are a consequence of the feature spaces capturing semantic characteristics that pertain most to the objective function used during training. FairFace is trained to classify perceived gender, which is correlated with hair length. On the other hand, Identity is trained to match faces corresponding to the same person, which should be invariant to hairstyle and hair length. SwAV is trained to match cropped views of an image, for which hair length is likely not a robust feature. Therefore, we suggest that generative model designers should not na\"ively expect in-domain feature spaces to be sensitive to all domain-specific semantics. Rather, we advocate carefully considering how the training objective may influence features, and empirically investigating these sensitivities.  

\textbf{Image-language models trained on massive general datasets capture many important semantic characteristics of faces.} The sensitivity analyses for both semantic attributes and distortions show the CLIP features are sensitive to all studied characteristics. In particular, CLIP provides a significant FD for all distorted facial region irrespective of the size of the region in pixels. This is likely because of two reasons: (1) CLIP is trained on a massive dataset, and (2) text provides a rich source of information on perceptual features to the image encoder that cannot have otherwise been learned using classical supervision. Based on these results, we encourage generative model designers to move away from perceptual features extracted from models trained on ImageNet (Inception, VGG~\cite{simonyanVeryDeep2015}, SwAV) and use large image-language models like CLIP. 

\textbf{StyleGAN2 consistently outperforms other face generators, except with respect to identities}. Table~\ref{tab:gen-model-eval}
shows that across almost all feature spaces and evaluation metrics, StyleGAN2 outperforms other face generators. However, when using Identity space, LDM and EG3D outperform StyleGAN2 in different ways. We speculate that because EG3D is trained to be aware of 3D geometry, it models features like the nose and eyes (which are important in Identity space as indicated by Figure~\ref{fig:causal-analysis-distortions2} well, yielding higher Precision. LDM may be able to capture a wider diversity of identities compared to GAN models because it is not susceptible to the mode collapse issues known to plague GANs~\cite{goodfellowGenerativeAdversarialNetworks2020}, yielding higher Recall and FD. This observation is further reinforced by the identity metrics for NVAE in comparison to EG3D. Although NVAE generates samples less precise in identity than StyleGAN2 and EG3D, it dethrones the GAN models for their third-place spots for Recall and FD. % \guha{I don't see really see this result that you wrote here in the table.}

\subsection{Limitations}

Our causal analysis of semantic attributes assumes perfectly counterfactual face pairs. However, it is not possible to exactly isolate one attribute from others when working with deep generators due to the correlations the generator learns from its training distribution. For example, when manipulating eyeglasses, we find that the skin texture becomes more wrinkly due to correlations between age and the propensity to wear eyeglasses. Moreover, in the case of synthesizing wrinkly faces, we find that the manipulation also causes the face to have squinted eyes. In general however, such correlations are known to be even more dramatic in real datasets~\cite{balakrishnanCausalBenchmarkingBiasin2021}, which makes synthetic generation a more attractive option for such analysis.

Our semantic attribute analysis uses a sample size of $1000$ images per set, which results in biased FD estimates~\cite{binkowskiDemystifyingMMDGANs2023, chongEffectivelyUnbiasedFID2020}. However, given that sample size was consistent throughout the experiment, the trend and shapes of curves shown in Figure~\ref{fig:causal-analysis-attrs} are accurate. This sample size bias is also a factor in our analysis for distortions of semantic regions such as hats and eyeglasses which are present in only a smaller subset($\sim10,000$) of the entire FFHQ dataset. %which may cause the FD for these regions to appear larger than their true estimate. 

We used publicly available deep networks for our feature spaces, but these networks vary in architecture type, size, and final layer feature number (for example, Inception and SwAV features have 2048 dimensions, while others have 512). These factors could be potential confounders in our causal analysis, though we believe that training set composition and objective functions likely have a far stronger influence on our results. 

% ----------------------------------------------------------- %
\section{Conclusion}
In this work, we proposed a strategy to causally evaluate the impact of variations in domain-specific characteristics on generative evaluation metrics using synthetic data in the context of face generation. We present a thorough study of the sensitivities FD computed using several deep feature spaces has with respect to facial attributes. Moreover, we provide an analysis of popular deep generative models evaluated in those feature spaces. The results of this study demonstrate that deep feature spaces have significant and unique biases over in-domain attributes due to both training data and objective functions. 

In summary, biases of a given feature space should be fully understood by researchers before using them for synthesis evaluation. Our experiments with face generators demonstrate the importance of considering multiple feature spaces during evaluation -- particularly those tuned to crucial details for the domain of interest (like identity for faces) -- to get a full picture of a model's strengths and shortcomings. As image generation models continue to improve at a rapid pace, such careful evaluation is necessary to make meaningful progress, including mitigating biases and enhancing the overall quality of generative image models.

\section{Broader impacts}
Improving the understanding of image generation evaluation metrics will enable researchers and developers to assess the performance of different generative models more effectively. This, in turn, promotes the development of more accurate, realistic, and reliable generation systems. Evaluation metrics also play a crucial role in mitigating biases of face generation models. Nevertheless, we recognize that face-generation technologies have far-reaching implications, including potentially harmful applications involving deepfakes and identity manipulation. Therefore, improving evaluation procedures of such methods can consequently increase the risks and harms associated with AI-generated facial content.

\bibliography{references}
\bibliographystyle{unsrt}

%%%%%%%%%%%%%%%%%%%%%%%%%%%%%%%%%%%%%%%%%%%%%%%%%%%%%%%%%%%%

\clearpage
\appendix
\counterwithin{figure}{section}
\counterwithin{table}{section}

\section{Implementation details}

\subsection{Open-source models}

We make use of many publically available, open-source codes, models, and parameters (checkpoints) for our work. Table~\ref{tab:model-urls} summarizes the models, code repositories, and checkpoint links used in our implementation. 

\begin{table}[ht]
\centering
\small
\caption{Summary of open-source codes, models, and parameters used in the implementation.}
\label{tab:model-urls}
\begin{tabular}{@{}p{1.25cm}p{1.75cm}p{3.5cm}p{5.5cm}@{}}
    \toprule
    \textbf{Model} & \textbf{Type} & \textbf{Code repository} & \textbf{Model checkpoint} \\ \midrule
    StyleGAN2 & FFHQ ($1024\times1024$) & \url{https://github.com/NVlabs/stylegan2-ada-pytorch} & \url{https://nvlabs-fi-cdn.nvidia.com/stylegan2-ada-pytorch/pretrained/ffhq.pkl} \\
    StyleCLIP & & \url{https://github.com/orpatashnik/StyleCLIP} & \\
    Stable Diffusion & v2 (Inpainting) & \url{https://github.com/huggingface/diffusers} & \url{https://huggingface.co/stabilityai/stable-diffusion-2-inpainting} \\
    Face segmentor & BiSeNet (CelebAMask-HQ) & \url{https://github.com/zllrunning/face-parsing.PyTorch} & \url{https://drive.google.com/open?id=154JgKpzCPW82qINcVieuPH3fZ2e0P812} \\
    InceptionV3 &  & \url{https://github.com/NVlabs/stylegan2-ada-pytorch} & \url{https://nvlabs-fi-cdn.nvidia.com/stylegan2-ada-pytorch/pretrained/metrics/inception-2015-12-05.pt} \\
    SwAV & ResNet-50 (800 epochs, batch size 4096) & \url{https://github.com/facebookresearch/swav} & \url{https://dl.fbaipublicfiles.com/deepcluster/swav_800ep_pretrain.pth.tar} \\
    CLIP & ViT-B/32 & \url{https://github.com/openai/CLIP} & \url{https://openaipublic.azureedge.net/clip/models/40d365715913c9da98579312b702a82c18be219cc2a73407c4526f58eba950af/ViT-B-32.pt} \\
    FairFace & ResNet-34 ~~ (7 race) & \url{https://github.com/dchen236/FairFace} & \url{https://drive.google.com/file/d/113QMzQzkBDmYMs9LwzvD-jxEZdBQ5J4X} \\
    SwAV-FFHQ & ResNet-50 (400 epochs, batch size 2048) & \url{https://github.com/facebookresearch/swav} & \url{https://storage.yandexcloud.net/yandex-research/ddpm-segmentation/models/swav_checkpoints/ffhq.pth} \\
    Identity & ResNet-34 (Glint360k) & \url{https://github.com/deepinsight/insightface} & \url{https://1drv.ms/u/s!AswpsDO2toNKq0lWY69vN58GR6mw?e=p9Ov5d} \\ 
    \bottomrule
\end{tabular}
\end{table}

\subsection{Counterfactual dataset: attributes}

\begin{table}[ht]
\centering
\small
\label{tab:cf-synthesis-methods}
\caption{Implementation parameters for our counterfactual attribute synthesis approach per semantic face attribute.}
\begin{tabular}{cccc}
    \toprule
    \multirow{2}{*}{\textbf{Attribute}} & \multirow{2}{*}{\textbf{Method}} & \multirow{2}{*}{\textbf{Text prompt}} & \textbf{Manipulation} \\
     &  &  & \textbf{parameters} \\
    \midrule
    \multirow{2}{*}{Hat} & Stable Diffusion & ``A photo of a face & Guidance \\
     & inpainting & with a hat" & scale $= 0.75$ \\
    \multirow{2}{*}{Eyeglasses} & \multirow{2}{*}{StyleCLIP} & Neutral: ``face" & $\alpha = 10$ \\
     &  & Target: ``face with eyeglasses" & $\beta= 0.13$ \\
    \multirow{2}{*}{Skin tone} & \multirow{2}{*}{OLLT} & \multirow{2}{*}{N/A} & Step \\
     &  &  & size $ =  0.5$ \\
    \multirow{2}{*}{Make-up} & \multirow{2}{*}{StyleCLIP} & Neutral: ``face" & $\alpha = 3$ \\
     &  & Target: ``face with makeup" & $\beta= 0.12$ \\
    \multirow{2}{*}{Wrinkly} & \multirow{2}{*}{StyleCLIP} & Neutral: ``face with skin" & $\alpha = 3$ \\
    &  & Target: ``face with wrinkly skin" & $\beta= 0.09$ \\
    \multirow{2}{*}{Smooth} & \multirow{2}{*}{StyleCLIP} & Neutral: ``face with skin" & $\alpha = -3$ \\
    &  & Target: ``face with wrinkly skin" & $\beta= 0.09$ \\
    \multirow{2}{*}{Chubby} & \multirow{2}{*}{StyleCLIP} & Neutral: ``face" & $\alpha = 5$ \\
    &  & Target: ``chubby face" & $\beta= 0.25$ \\
    \multirow{2}{*}{Slim} & \multirow{2}{*}{StyleCLIP} & Neutral: ``face" & $\alpha = -5$ \\
    &  & Target: ``chubby face" & $\beta= 0.25$ \\
    \multirow{2}{*}{Slim} & \multirow{2}{*}{StyleCLIP} & Neutral: ``face" & $\alpha = -5$ \\
    &  & Target: ``chubby face" & $\beta= 0.25$ \\
    \multirow{2}{*}{Frowning} & \multirow{2}{*}{StyleCLIP} & Neutral: ``smiling face" & $\alpha = 5$ \\
    &  & Target: ``frowning face" & $\beta= 0.20$ \\
    \multirow{2}{*}{Hair length} & \multirow{2}{*}{StyleCLIP} & Neutral: ``face with hair" & $\alpha = 15$ \\
    &  & Target: `face with long hair" & $\beta= 0.20$ \\
    \multirow{2}{*}{Curly} & \multirow{2}{*}{StyleCLIP} & Neutral: ``face with hair" & $\alpha = 5$ \\
    &  & Target: `face with curly hair" & $\beta= 0.25$ \\
    \multirow{2}{*}{Fringe} & \multirow{2}{*}{StyleCLIP} & Neutral: ``face with hair" & $\alpha = 5$ \\
    &  & Target: `face with fringe hair" & $\beta= 0.15$ \\
    \bottomrule
\end{tabular}
\end{table}

We use a two-step process to create our counterfactual facial attribute dataset. We first synthesize a set of \emph{base faces} that exhibit predefined uniform characteristics of light skin tones and short hair, and no: facial hair, make-up, frowning expressions, hats, or eyeglasses. To accomplish this, we sample a set of intermediate-style latent vectors $\{\mathbf{w_i}: \mathbf{w_i} \in \mathcal{W}\}$. We then use orthogonalized linear latent space traversals (OLLT) to traverse the latent vectors in a direction corresponding to light skin tone and short hair\footnote{The hyperplane coefficients for age, gender, hair length, and skin tone attributes were graciously provided by the authors upon request.}. Finally, we filter these faces via human evaluations to ensure they meet the defined criteria. The final number of base faces contained in the dataset amounted to $1427$ images.  

In the second step, we synthesize counterfactual pairs from the base faces for each of the 12 binary attributes analyzed in our experiments (see first column of Table~\ref{tab:cf-synthesis-methods}). To accomplish this, we utilize one of three different image manipulation methods based on the attribute type: (1) OLLT, (2) StyleCLIP~\cite{patashnikStyleCLIPTextDrivenManipulation2021}, and (3) image inpainting with Stable Diffusion~\cite{vonplatenDiffusersStateoftheartDiffusion2023}. We choose the best method for each attribute based on a qualitative assessment of how well each method can manipulate the attribute while holding others constant. A summary of the manipulation method and parameters used for each attribute is listed in Table~\ref{tab:cf-synthesis-methods}. To manipulate skin tone, we use OLLT to traverse in the direction of dark skin tones. For wearing a hat, we first automatically mask out a region reaching from the bottom of the forehead to the top of the image using 3D facial landmarks detected by MediaPipe face mesh model~\cite{lugaresiMediaPipeFrameworkBuilding2019}. We then performed image inpainting using Stable Diffusion with the prompt ``a photo of a face with a hat". For all other attributes, we use StyleCLIP to traverse along a direction that corresponds to the text prompts detailed in Table~\ref{tab:cf-synthesis-methods}. Note that for some attributes, namely ``slim" and ``smooth", we traverse in the negative direction of the text prompt. We experimentally found that these attributes are best manipulated by traversing in these negative directions as opposed to the corresponding positive directions (e.g. ``slim face" and ``face with smooth skin").    

\subsection{Counterfactual dataset: distortions (blur)} 

To create our counterfactual distortions (blur) dataset, we apply heavy blur to 9 semantic regions on real FFHQ face images. The regions for each image are obtained using segmentation masks obtained from a public face segmentation model (see Table~\ref{tab:model-urls}). The heavy blur is defined as a Gaussian blur with kernel size of $111\times111$ pixels and standard deviation $\sigma = 100$ pixels applied to a $512\times512$ image. The counterfactuals are synthesized by replacing the region of interest in the real image with the corresponding region in the blurred image. 

% ----------------------------------------------------------- %

\section{Additional experimental results}

\begin{figure}[b]
    \centering
    \includegraphics[width=0.85\linewidth]{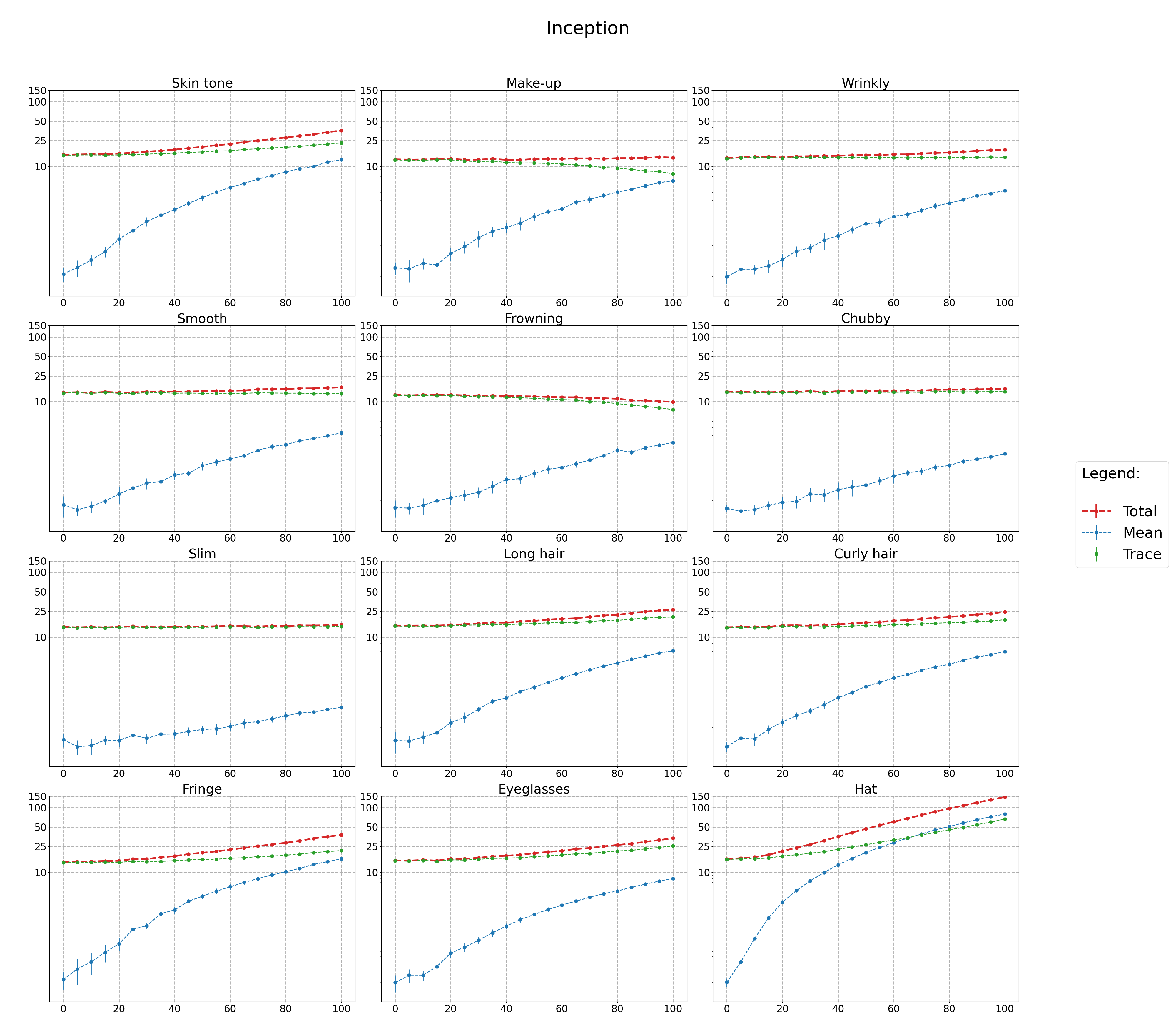}
    \caption{Results for causal sensitivity analysis of Fr\'echet distances in the Inception feature space.}
    \label{fig:frechet-inception-component-breakdown}
\end{figure}

\begin{figure}
    \centering
    \includegraphics[width=0.85\linewidth]{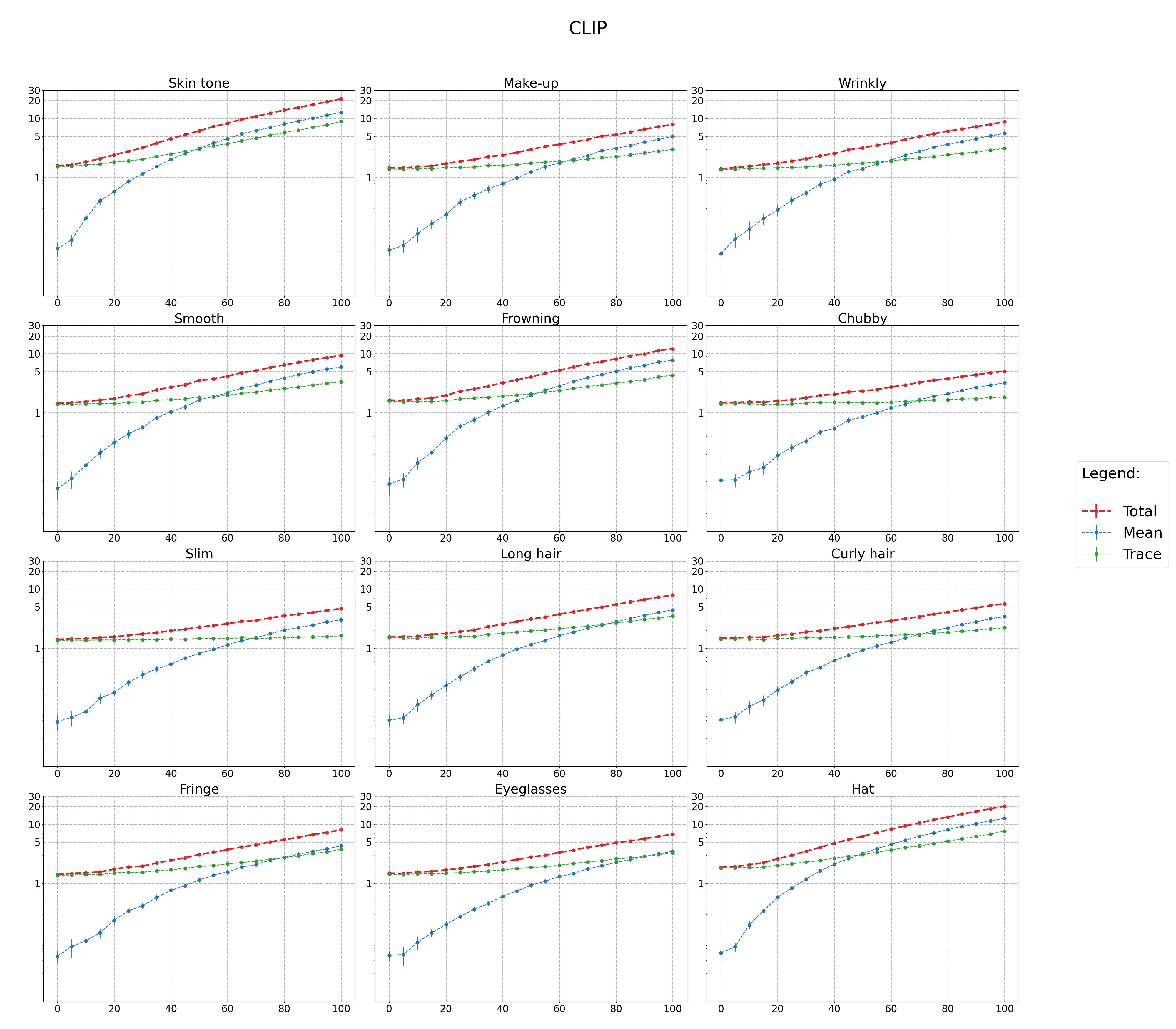}
    \caption{Results for causal sensitivity analysis of Fr\'echet distances in the CLIP feature space.}
    \label{fig:frechet-clip-component-breakdown}
\end{figure}

\begin{figure}
    \centering
    \includegraphics[width=0.85\linewidth]{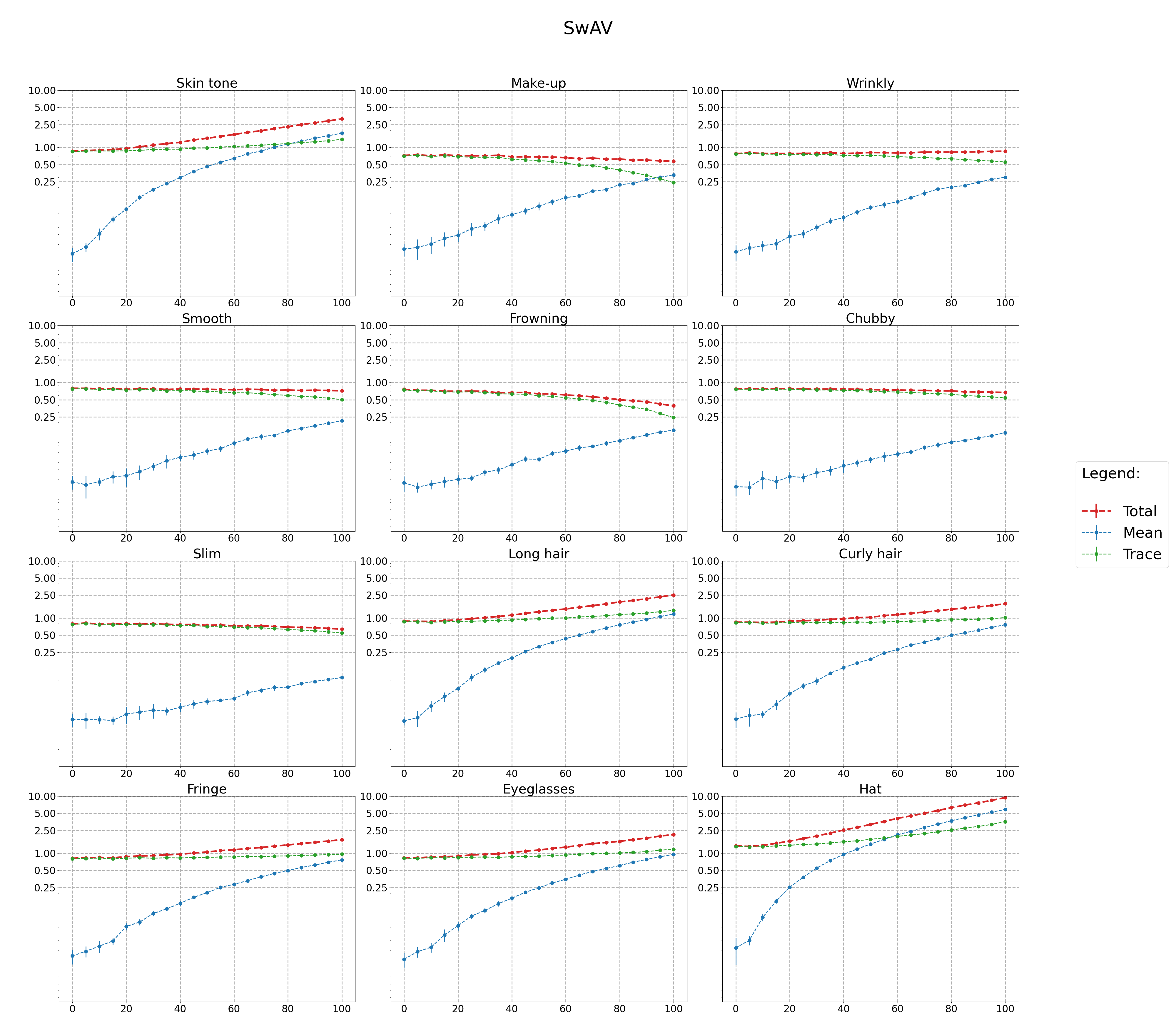}
    \caption{Results for causal sensitivity analysis of Fr\'echet distances in the SwAV feature space.}
    \label{fig:frechet-swav-component-breakdown}
\end{figure}

\begin{figure}
    \centering
    \includegraphics[width=0.85\linewidth]{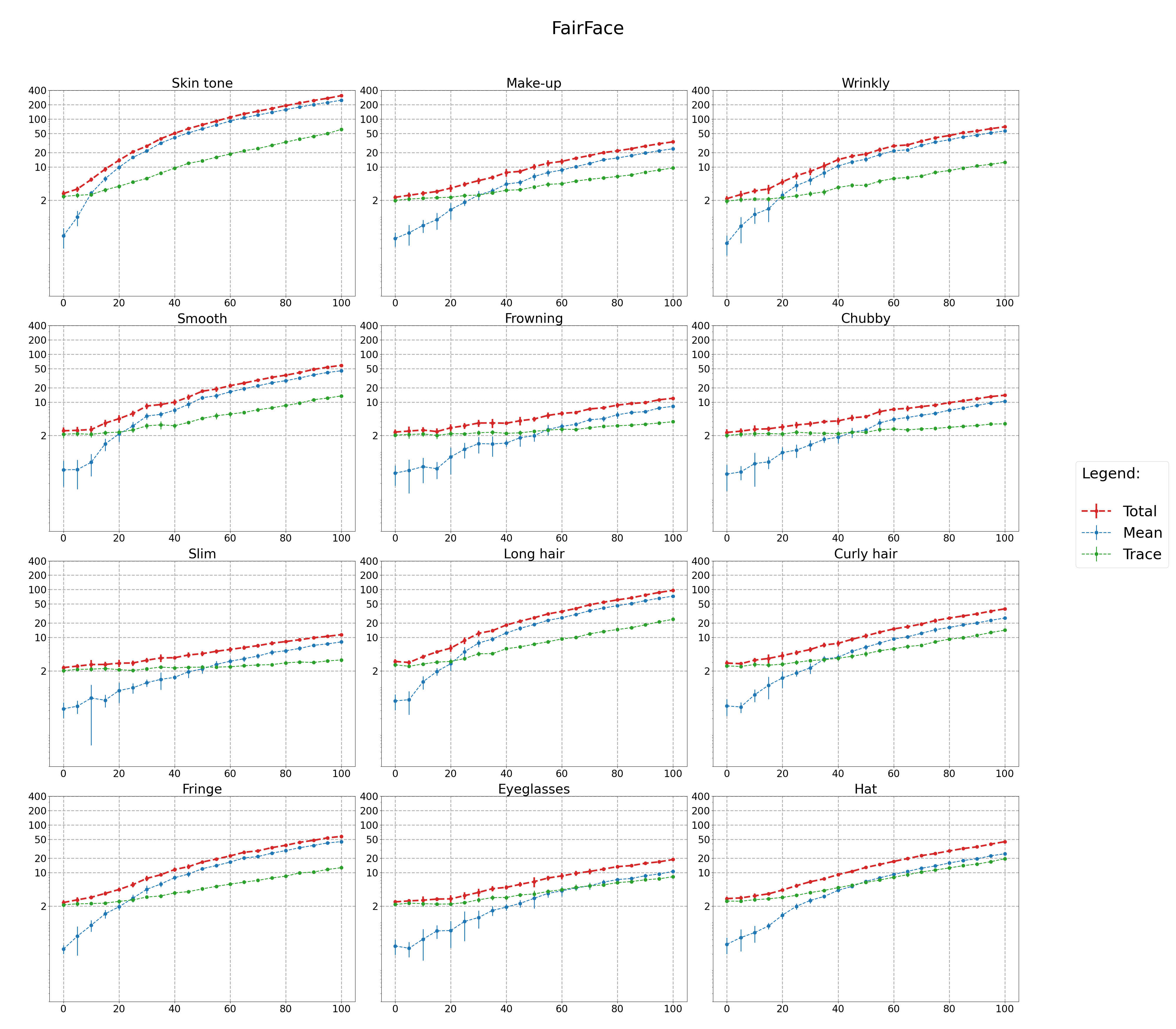}
    \caption{Results for causal sensitivity analysis of Fr\'echet distances in the FairFace feature space.}
    \label{fig:frechet-fairface-component-breakdown}
\end{figure}

\begin{figure}
    \centering
    \includegraphics[width=0.85\linewidth]{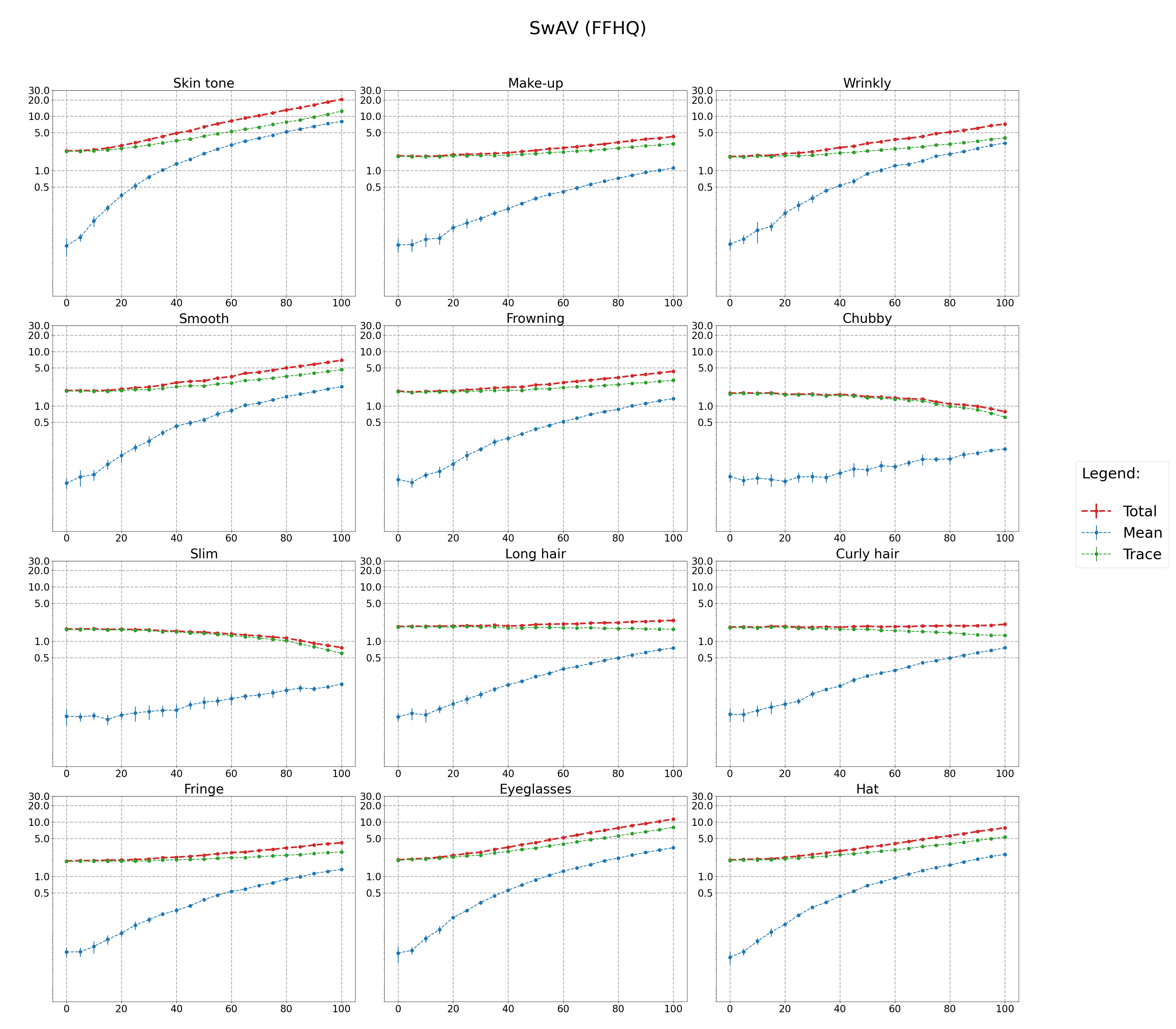}
    \caption{Results for causal sensitivity analysis of Fr\'echet distances in the SwAV-FFHQ feature space.}
    \label{fig:frechet-swav-ffhq-component-breakdown}
\end{figure}

\begin{figure}
    \centering
    \includegraphics[width=0.85\linewidth]{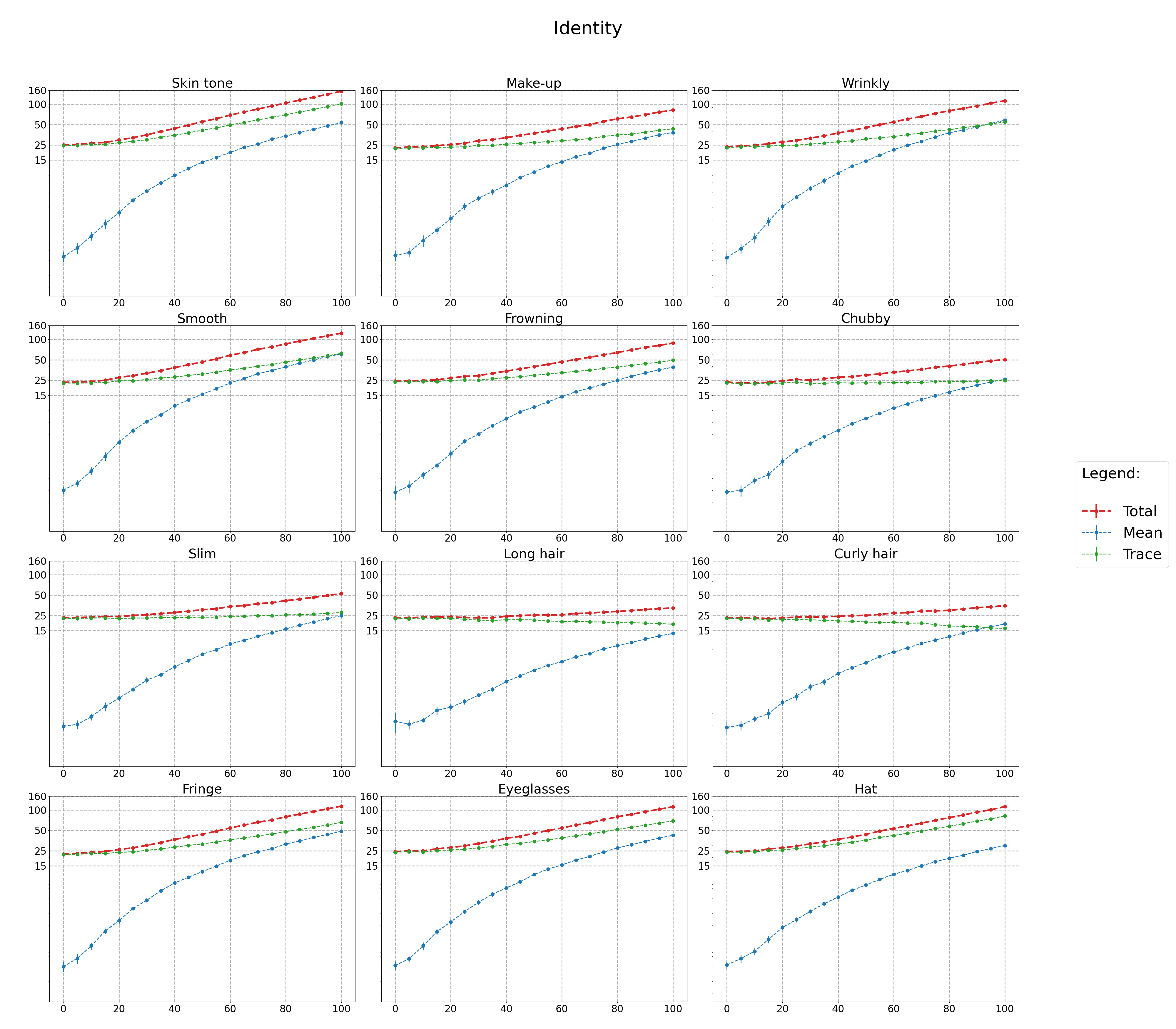}
    \caption{Results for causal sensitivity analysis of Fr\'echet distances in the identity feature space.}
    \label{fig:frechet-identity-component-breakdown}
\end{figure}

\begin{figure}
  \label{fig:causal-analysis-distortions1}
  \centering
  \includegraphics[width=\linewidth]{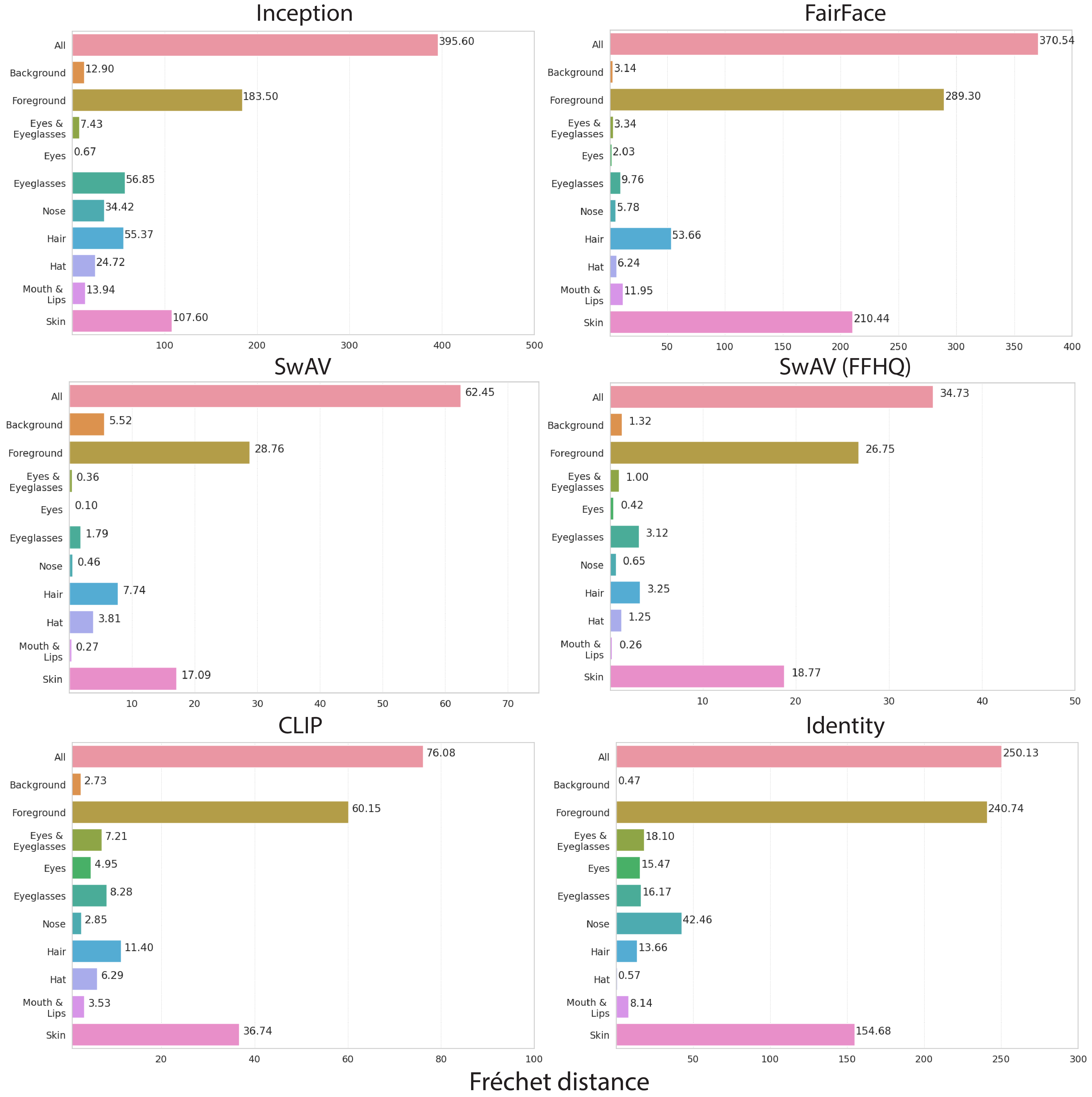}
  \caption{Results for the effect of semantic region distortion (blur) on Fr\'echet distances (FD) across different feature spaces.}
\end{figure}

In this section, we present the full results for the causal sensitivity analyses of Fr\'echet distance (FD) in all 6 feature spaces to image characteristics. Figures~\ref{fig:frechet-inception-component-breakdown} to \ref{fig:frechet-identity-component-breakdown} plot the FD against differences in facial attribute proportions across all 12 attributes analyzed. Additionally, the mean and trace terms that contribute to the total FD are shown. Figure~\ref{fig:causal-analysis-distortions1} plots the (unnormalized) FD for each feature space when the specified semantic region is heavily blurred. 

\end{document}